\documentclass[letterpaper, 10 pt, conference]{ieeeconf}  

\IEEEoverridecommandlockouts                              

\usepackage{graphicx} 
\usepackage{amsmath} 
\usepackage{siunitx}
\usepackage{booktabs}
\usepackage{caption}
\usepackage{subcaption}
\usepackage{float}
\usepackage{placeins}
\usepackage{hyperref}

\title{\LARGE \bf
Advanced Skills by Learning Locomotion and Local Navigation End-to-End
}

\author{Nikita Rudin$^{1, 2}$, David Hoeller$^{1, 2}$, Marko Bjelonic$^{1}$ and Marco Hutter$^{1}$
\thanks{$^{1}$ Robotic Systems Lab, ETH Zurich. $^{2}$ NVIDIA, contact: rudinn@ethz.ch. \newline This work is supported in part by the EU Horizon 2020 programme grant agreement No.852044}
}

\begin{document}

\maketitle
\thispagestyle{empty}
\pagestyle{empty}

\begin{abstract}
The common approach for local navigation on challenging environments with legged robots requires path planning, path following and locomotion, which usually requires a locomotion control policy that accurately tracks a commanded velocity. However, by breaking down the navigation problem into these sub-tasks, we limit the robot's capabilities since the individual tasks do not consider the full solution space.
In this work, we propose to solve the complete problem by training an end-to-end policy with deep reinforcement learning.
Instead of continuously tracking a precomputed path, the robot needs to reach a target position within a provided time. The task's success is only evaluated at the end of an episode, meaning that the policy does not need to reach the target as fast as possible. It is free to select its path and the locomotion gait. 
Training a policy in this way opens up a larger set of possible solutions, which allows the robot to learn more complex behaviors.
We compare our approach to velocity tracking and additionally show that the time dependence of the task reward is critical to successfully learn these new behaviors. Finally, we demonstrate the successful deployment of policies on a real quadrupedal robot.
The robot is able to cross challenging terrains, which were not possible previously, while using a more energy-efficient gait and achieving a higher success rate. Supplementary videos can be found on the project website: \url{https://sites.google.com/leggedrobotics.com/end-to-end-loco-navigation}
\end{abstract}

\section{INTRODUCTION}
\label{sec:introduction}

\begin{figure}[!tb]
    \centering
    \begin{subfigure}[b]{0.6\columnwidth}
        \centering
        \includegraphics[width=\textwidth, trim={0mm, 0mm, 0mm, 0mm}, clip]{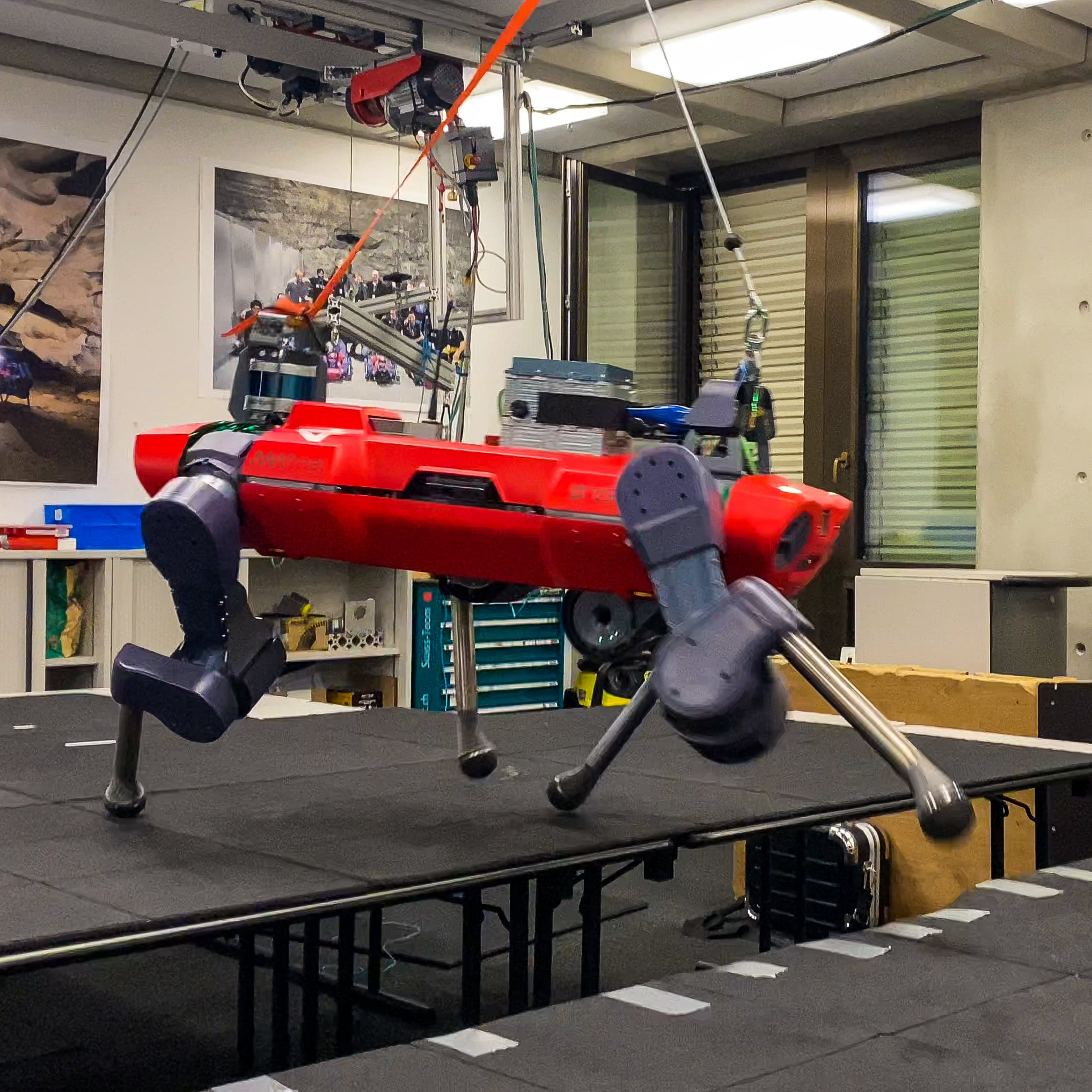}
     \end{subfigure}
    \begin{subfigure}[b]{0.3\columnwidth}
    \centering
        \includegraphics[width=\textwidth, trim={0mm, 0, 0mm, 0}, clip]{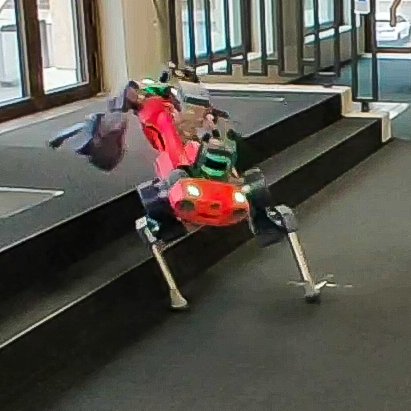}
        \includegraphics[width=\textwidth, trim={0mm, 0, 0mm, 0}, clip]{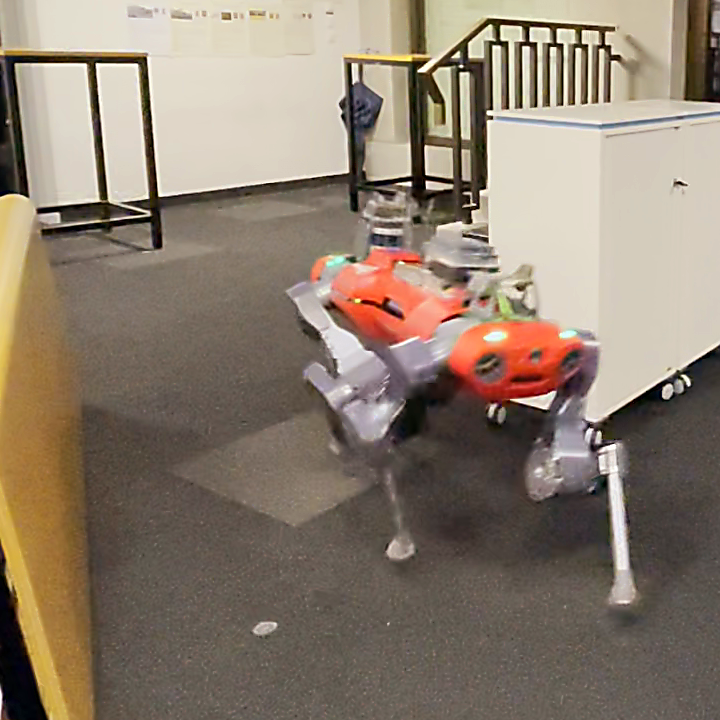}
     \end{subfigure}

    \caption{The legged robot ANYmal navigating around obstacles, climbing stairs, and jumping over a gap.}
    \label{fig:intro}
    \vspace{-5mm}
\end{figure}

Legged robots have the potential to navigate challenging environments, such as in Fig. \ref{fig:intro}, much better than wheeled robots, but they require complex motions to do so. Multiple control methods have been proposed including Model Predictive Control \cite{KimMPC, NeunertWholeBody}, trajectory optimization \cite{apgar2018fast, BjelonictrajOpt}, imitation learning \cite{RoboImitationPeng20, ReskeMpcnet}, and fully learned approaches \cite{leeLearning, CassieBlind}. Recently, deep reinforcement learning has proven to be a powerful tool to achieve robust locomotion. It has brought the robots out of a controlled lab environment and allowed them to evolve in the real world, using perceptive and proprioceptive information to walk on obstacles such as stairs, rocks, or roots \cite{miki2022learning}.

To achieve a robust execution in the real world, the controller must be capable of dynamic motions while handling model mismatches, noisy sensors, reacting to external disturbances, missteps, and slippage. Nonetheless, these policies still fail to exhibit the level of agility and precision of animals or humans. In all of these cases, the control policy only needs to slightly modulate a periodic motion of the joints. In contrast, animals can switch gaits, plan a series of consecutive steps, perform jumps, and use contacts on other parts of their body when necessary. 
If we want our policies to achieve the same level of dexterity, they need to explore the entire solution space by removing assumptions and implicit constraints that limit the possible behaviors.

\subsection{Related Work}

Most of the related work splits the local navigation task of legged robots into multiple sub-parts. For example, a planing module generates a trajectory that the robot can follow to reach its target. Then a path-following controller steers the robot towards that path by providing velocity commands to the locomotion controller, which tries to match those commands by moving the legs of the robot \cite{BigDogNavigation, anymalNavigation}. Similarly, when operators manually guide the robot through a local environment with a joystick, they send velocity commands to the locomotion controller. As such, velocity tracking is commonly used as the task for both model-based \cite{KimMPC, NeunertWholeBody, GuoCassieWholeBody} and learned locomotion controllers \cite{leeLearning, CassieBlind, miki2022learning, jilearnedEstimator}. 

This is a natural way to split the locomotion and navigation problems. Unfortunately, it imposes unintended restrictions on the behavior of the low-level policy and thus limits the possibilities of the overall system. For example, when jumping over a gap, the robot needs to be free to modulate its velocity before and after the jump. Furthermore, the locomotion policy must be allowed to slow down or even stop in dangerous situations. These cases would result in higher velocity tracking costs making the desired behaviors harder to achieve. Finally, keeping the base at a constant velocity encourages the policy to learn a gait, which is not necessarily the most energy-efficient. In the case of four-legged locomotion, this results in the trotting gait seen in all learning approaches mentioned above. Even a policy trained specifically to generate gaits, only learns to output different versions of trotting \cite{gaitAdaptation}.

Trajectory optimization methods can directly produce joint commands based only on a target state and a cost to minimize \cite{apgar2018fast, anymalTrajOpt}. Unfortunately, these methods are computationally expensive and lack the robustness of reinforcement learning policies.
A simple position tracking task would penalize the distance between the robot and the target location continuously. However, the optimal solution would be to reach the goal as fast as possible, resulting in a behavior close to the "bang-bang" solution of optimal control problems \cite{bellman1956bang}. This is not the desired behavior for a real-world robot as it would result in needlessly aggressive and dangerous motions. This problem can be mitigated by penalizing velocities and torques, but the final behavior depends on the relative scale of these penalties, which must be tuned individually for each case. Additionally, this set-up also penalizes trajectories where the robot slows down or deviates from the shortest path.

Previous works have explored the agile locomotion of legged robots. Trajectory optimization has been shown to produce agile motions, including jumping and climbing behavior \cite{anymalTrajOpt}. However, this approach requires costly optimizations for each trajectory, assumes the model to be continuously differentiable, and the authors only show hardware experiments on simple cases. Learning approaches, splitting the locomotion problem into a foothold planner and a tracker \cite{deepGait, visLocomotion}, achieved locomotion over challenging terrains, including gaps and stepping stones. Despite impressive results, this approach requires a multi-stage training approach, assumes contact patterns that would fail on our climbing task, and was again not shown to transfer to the real world. For selected tasks, multiple approaches resulted in dynamic and agile motions on real robots. These include: jumping over obstacles with a model-based planer \cite{ParkJumping}, jumping onto a table with trajectory optimization \cite{NguyenOptimizedJump}, exhibiting cat-like landing capabilities with a model-based approach \cite{jeonLandingOpt}, a learned policy \cite{rudin2021cat}, as well as a combination of both \cite{kurtzFallingCat}.

Learning approaches have been applied to local navigation of a legged robot. The problem can be split into a high-level policy steering the low-level locomotion controller \cite{hoellerObsAvoidance}, or combined into a single end-to-end policy \cite{visionGuidedQL}. However, in both cases the locomotion controllers are trained using velocity commands resulting in limited capabilities of the system. 

\subsection{Contribution}

In this work, we propose a position-based formulation where the robot must reach the target location only after a defined time. The task reward is based on the final distance to the target at the end of an episode and is not influenced by the trajectory the robot took to get there. We need to add multiple penalties to achieve sim-to-real transfer, but this setup has the advantage of decoupling the main task from these costs. As long as the robot reaches the target position, it is free to optimize its path and gait to minimize the additional penalties without influencing the main task reward. This creates a hierarchical reward structure where the policy first learns to solve the task before optimizing the quality of motion needed to do so. 

Using the velocity tracking and continuous position approaches as baselines for our experiments, we show that our method allows the policy to succeeds on common terrains such as stairs and slopes with increased difficulty. Furthermore, it enables the training complex behaviors such as jumping over gaps, climbing on boxes, and avoiding obstacles. Additionally, the policy learns a novel locomotion gait resulting in lower failure rates and energy consumption. Finally, our approach removes the need for a local planner since the locomotion policy directly learns this capability. The policies are deployed on the real robot and run in real-time on the onboard computer. 


\section{METHOD}
We train a neural network policy to solve various locomotion tasks for a quadrupedal robot. The policy receives measurements provided by the robot's sensors, and outputs commands that are directly sent to the motors. The neural network is trained in simulation using Deep RL, after which it is transferred to the real robot without any further training or adaptation. We build upon the massively parallel set-up of \cite{RudinMinutes}, with Isaac Gym \cite{IsaacGym} simulation and Proximal Policy Optimization (PPO) \cite{PPO} as the Deep RL algorithm.


\subsection{Observation and Action Space}
The observations include joint positions, joint velocities, base linear and angular velocity, commands, previous actions, and terrain measurements sampled around the robot. The commands are defined as the three-dimensional location of the target expressed in the base frame and the remaining time to reach that location. Noise is added to the observations based on the noise level of the robot's sensors.
The actions are interpreted as target joint positions given to the motors where a PD controller transforms them into joint torques.

\subsection{Rewards}
The rewards are split into two parts: task rewards only active at the end of the episode, and various penalties provided throughout the episode. 
\subsubsection{Task Reward} 
The task reward instructs the robot to reach the provided target location. It is calculated as
\begin{equation}
r_{task}=
\begin{cases}
    \frac{1}{T_r}\frac{1}{1+\|\mathbf{x}_b-\mathbf{x}_b^*\|^2}, & \text{if } t> T - T_r\\
    0, & \text{otherwise}.
 \end{cases}
\end{equation}
where $\mathbf{x}_b$ and $\mathbf{x}_b^*$ are the current and target positions of the base, $t$ is the time since the beginning of the episode, $T$ is the maximum episode length and $T_r$ is a parameter defining the duration of the reward. $T_r$ needs to be long enough to force the policy to stop at the target in a stable configuration. Otherwise, the policy learns to jump or lean towards the target at the last moment, which would result in a crash on the real robot where there are no episode resets.

\subsubsection{Penalties} 
\label{sec:penalties}
We penalize joint accelerations, joint torques, collisions, and abrupt actions changes. Additionally, we introduce a new reward penalizing feet accelerations. It reduces impacts with the ground, causing loud noises and potential damage to the motors and structure. The complete set of penalties is computed as
\begin{equation}
\begin{split}
r_{penalties} =& -c_1\|\mathbf{\ddot{q}}\|^2 - c_2\|\boldsymbol{\tau}\|^2 - c_3 N_c \\ 
& -c_4\|\mathbf{a} - \mathbf{a}_{-1}\|^2 - c_5 \sum_{feet}\|\mathbf{\ddot{x}}_f\|^2
\end{split}
\end{equation}
where $\mathbf{q}$, $\boldsymbol{\tau}$, $N_c$ $\mathbf{a}$, and $\mathbf{x}_f$ represent joint positions, joint torques, number of collisions, actions, and positions of the feet respectively, and $c_{1-5}$ are scaling constants.

There is an important difference in how these penalties interact with the main task. With velocity tracking, the policy must compromise between minimizing these penalties and optimizing the main task. On the other hand, minimizing these penalties does not counteract the main reward with our approach. As long as the robot reaches the target in time, it can freely minimize the different quantities without negatively affecting the position tracking. This simplifies the tuning process and removes the need to compromise on the quality of the solution.

\subsubsection{Exploration Reward}
Due to the sparse nature of the main reward, we find it beneficial to add a reward term encouraging exploration at the beginning of training. In order to bias the policy to walk towards the target, we reward it for any base velocity in the correct direction. The reward is defined as 
\begin{equation}
r_{bias} = \frac{\mathbf{\dot{x}}_{b} \cdot (\mathbf{x}_b^* - \mathbf{x}_b)}
{\|\mathbf{\dot{x}}_{b}\|\|\mathbf{x}_b^* - \mathbf{x_b}\|}
\end{equation}

This rewards is automatically removed after a few training iterations, and thus doesn't constrain the final solution learned by the policy. $r_{bias}$ is removed once $r_{task}$ reaches \SI{50}{\%} of it's maximum value. 

\subsubsection{Stalling Penalty}
PPO optimizes the discounted sum of rewards over the episode, with a discount factor that must be lower than 1 for stability. This has an unintended effect on the optimal solution to the problem. The policy receives two types of rewards, a positive reward at the end of an episode representing the main task and negative penalties throughout the episode, which are larger when the robot is walking. In order to optimize the discounted sum of rewards, it is beneficial to push the negative penalties as far into the future as possible. Therefore, the policy learns to wait until the last moment when it suddenly runs fast towards the target. In practice, we would like the robot to either use all of the provided time or wait at the target. Furthermore, the real-world energy consumption should be optimized without a discount factor.
In order to counterbalance this effect, we add a small penalty waiting while being far away from the target. It is defined as
\begin{equation}
r_{stall} = 
\begin{cases}
    -1, & \text{if}\quad \|\mathbf{\dot{x}}_b\| < \SI{0.1}{m/s}\\
    &\text{and } \|\mathbf{x}_b - \mathbf{x}_b^*\| > \SI{0.5}{m}\\
    0, & \text{otherwise}.
 \end{cases}
\end{equation}

\subsection{Terrains}
\begin{figure}[!tb]
    \centering
    \includegraphics[width=\columnwidth]{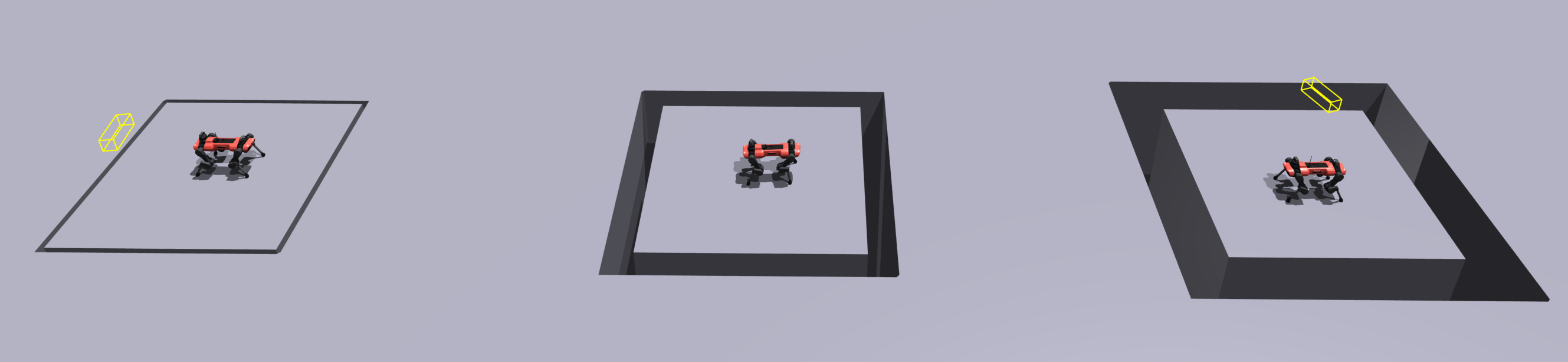}
    
    \vspace{1mm}
    
    \includegraphics[width=\columnwidth]{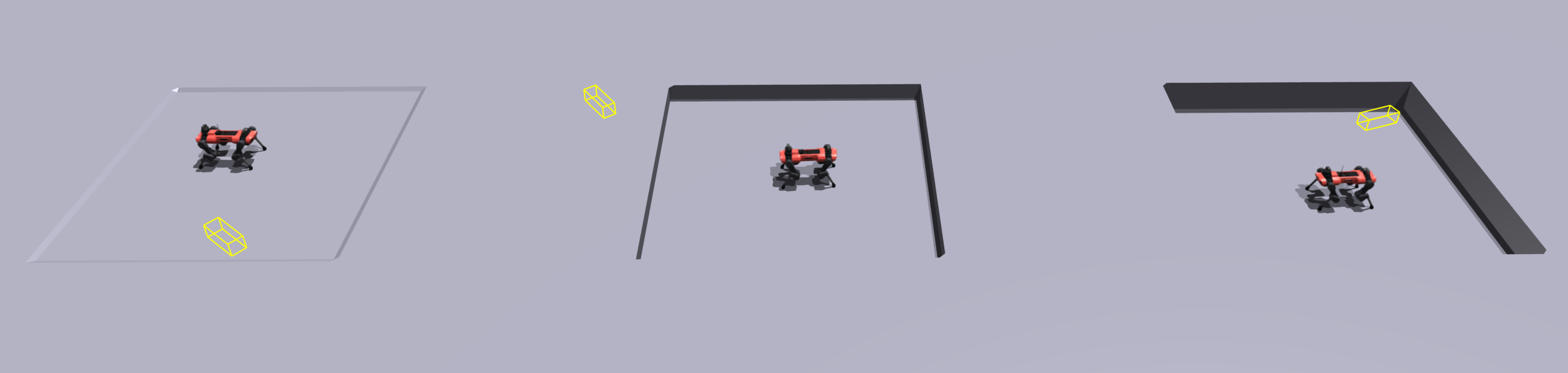}
    
    \vspace{1mm}
    
    \includegraphics[width=\columnwidth]{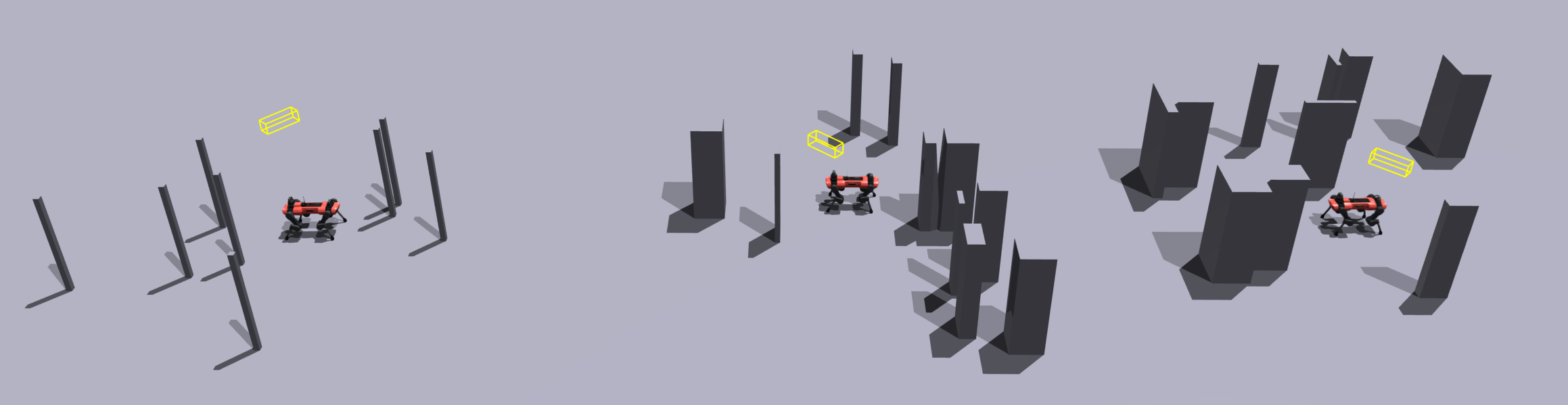}

    \caption{Examples of three newly added terrains types with a representative curriculum from the easiest version on the left to the hardest terrain seen during training on the right.}
    \label{fig:new_terrains}
\end{figure}
For a proper comparison with velocity tracking, we first train a policy on terrains from \cite{RudinMinutes}. These include slopes, stairs, random rectangular obstacles, and rough surfaces. We then explore terrains requiring more agile and complex maneuvers. We evaluate our approach on three new tasks represented in Fig. \ref{fig:new_terrains}. First we create a terrain where the policy needs to step or jump over a gap. The size of the gap is such that keeping a constant base velocity is too constraining since the robot needs to leap forward to cross the distance and inevitably slows down after the jump side since it barely reaches the other side. Second, we explore the use of other parts of the body by creating a climbing task. In this scenario the robot must climb out of a pit, which is too deep to simply step out of it. The policy needs to learn a complex series of actions where it first puts the front feet up before using the back knees to help it reach the top. Finally, we test the local planning capabilities of our policy by creating a terrain with high obstacles which the policy must avoid. When an obstacle is in its way, the policy must learn to disregard the direct path towards the target and find a way around the obstacle.

\subsection{Episode Process}
At the beginning of each episode, a valid position target is sampled for each robot. The sampling is done uniformly in polar coordinates around the initial position of the robot with a radius between \SI{1}{m} and \SI{5}{m}. The height of the target is set to \SI{0.5}{m} above the ground. A target is considered invalid if it overlaps with holes or high obstacles. Invalid targets are re-sampled until none are left.
The episode length is set to $T=\SI{6}{s}$, with the task reward provided after \SI{5}{s} for a duration of $T_r=\SI{1}{s}$.
At the end of the episode, the progress of each robot is evaluated to update the terrain curriculum. A terrain is considered successfully solved if the robot finishes the episode within \SI{0.5}{m} of the target.

\subsection{Training Stability}
We have found that while our approach can lead to better-performing policies, it is also more sensitive to training instabilities. In particular, predicting the \textit{value function} seems to be more difficult with our temporally sparse task reward. The fact that the task reward is only provided at the end of the episode and only if the robot does not crash until then makes it harder for the \textit{critic} to predict the correct outcome. This makes the whole system prone to instabilities, where a random seed change can cause a failed training run.
We use the PPO implementation from \cite{RudinMinutes} with similar hyper-parameters. However, we introduce some modifications which have proven effective to mitigate the instabilities described above.

First, we increase the batch size by doubling the number of consecutive steps each robot takes per policy iteration. We use 4096 parallel robots performing 48 steps resulting in a \textit{batch size} of $~200'000$ samples. Additionally, we reduce the episode length from \SI{20}{s} to \SI{6}{s}. Finally, the \textit{value function bootstrapping} was introduced to mimic an infinite horizon problem, which is necessary in the case of velocity tracking since there is no termination condition for this task. However, in our case, the time until task termination is clearly defined and provided as input to both the \textit{actor} and the \textit{critic}. We, therefore, remove the bootstrapping and obtain more stable training.
We train all policies for a total of 2000 iterations requiring approximately 1 hour of training.

\section{Simulation Experiments}
In this section, we compare our approach to the velocity tracking task and a continuous position tracking reward. The velocity tracking policy is trained as in \cite{RudinMinutes} with the only difference being reward scales and command ranges. We use the penalties described in Sec. \ref{sec:penalties} with the same weights while scaling the task rewards to a magnitude similar to the tracking reward of the new approach. In order to simplify the task, we remove sideways velocity commands keeping only forward/backward and yaw.
Continuous position tracking is achieved by providing $r_{task}$ at every time-step ($T_r = T$) without any further tuning.
\subsection{Maximum Terrain Difficulty}
\begin{figure}[!tb]
    \centering
    \begin{subfigure}[b]{0.32\columnwidth}
        \centering
        \includegraphics[width=\textwidth, trim={0mm, 0, 0mm, 0}, clip]{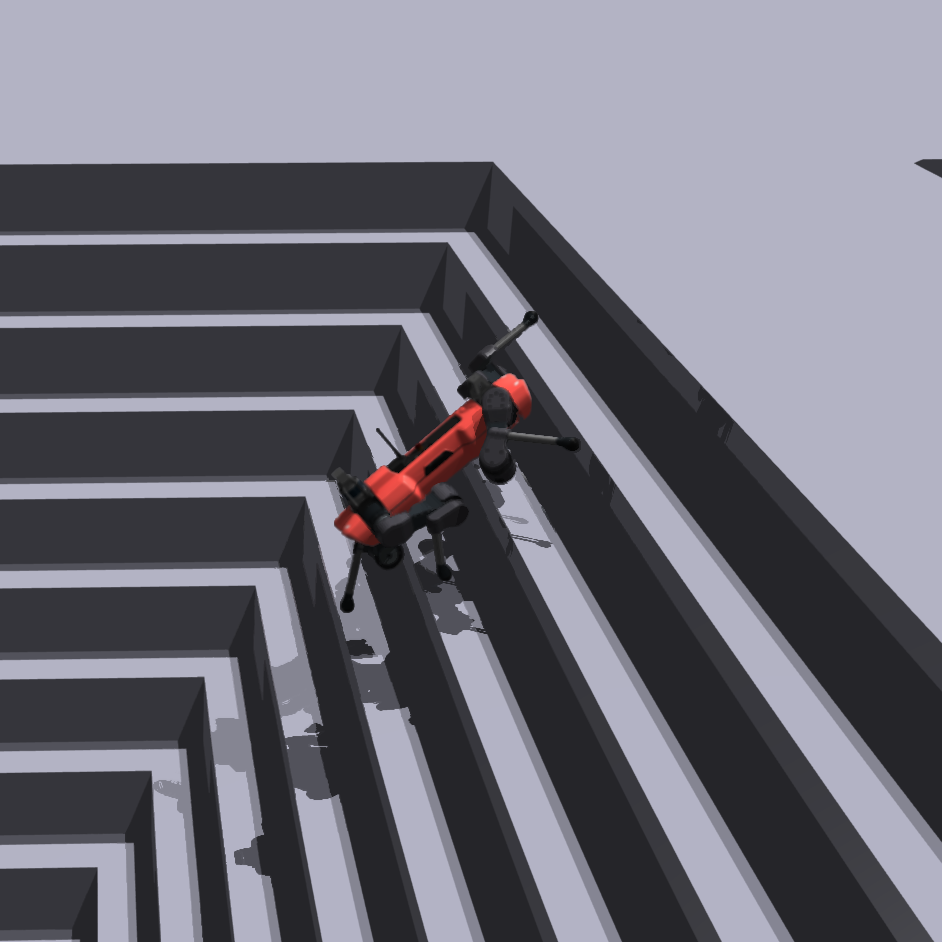}
     \end{subfigure}
     \hfill    
     \begin{subfigure}[b]{0.32\columnwidth}
     \centering
        \includegraphics[width=\textwidth, trim={0mm, 0, 0mm, 0}, clip]{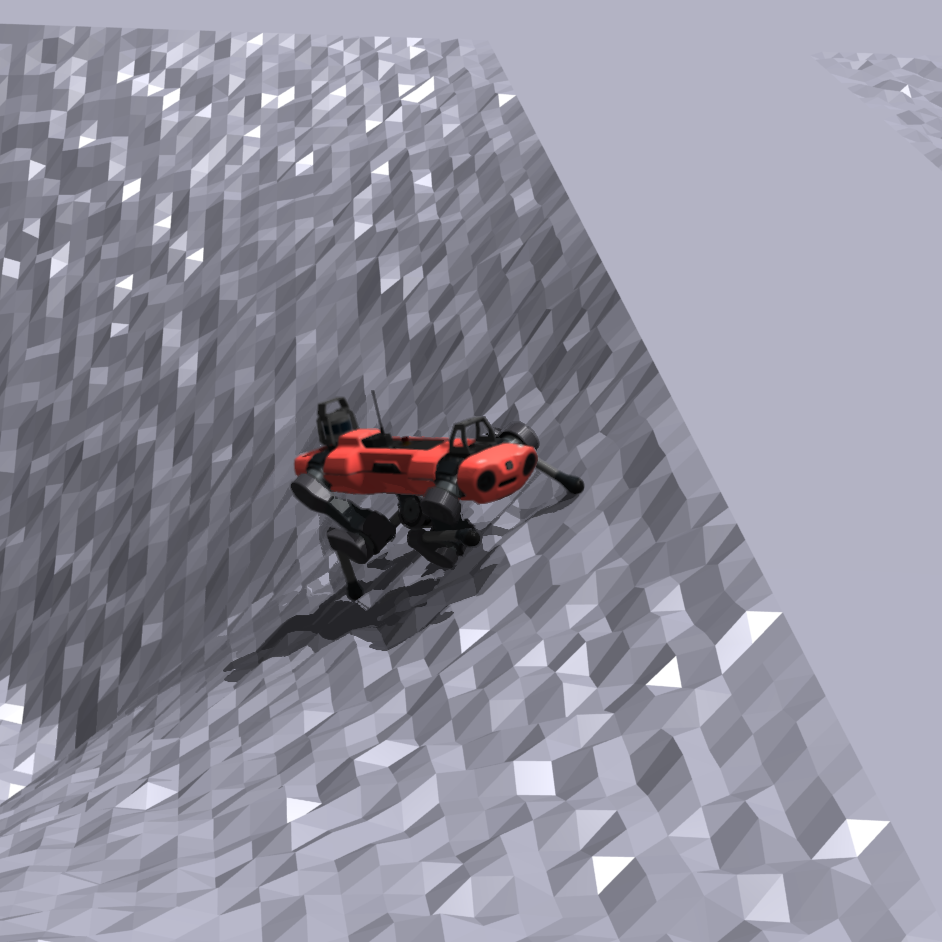}
     \end{subfigure}
     \hfill
    \begin{subfigure}[b]{0.32\columnwidth}
    \centering
        \includegraphics[width=\textwidth, trim={0mm, 0, 0mm, 0}, clip]{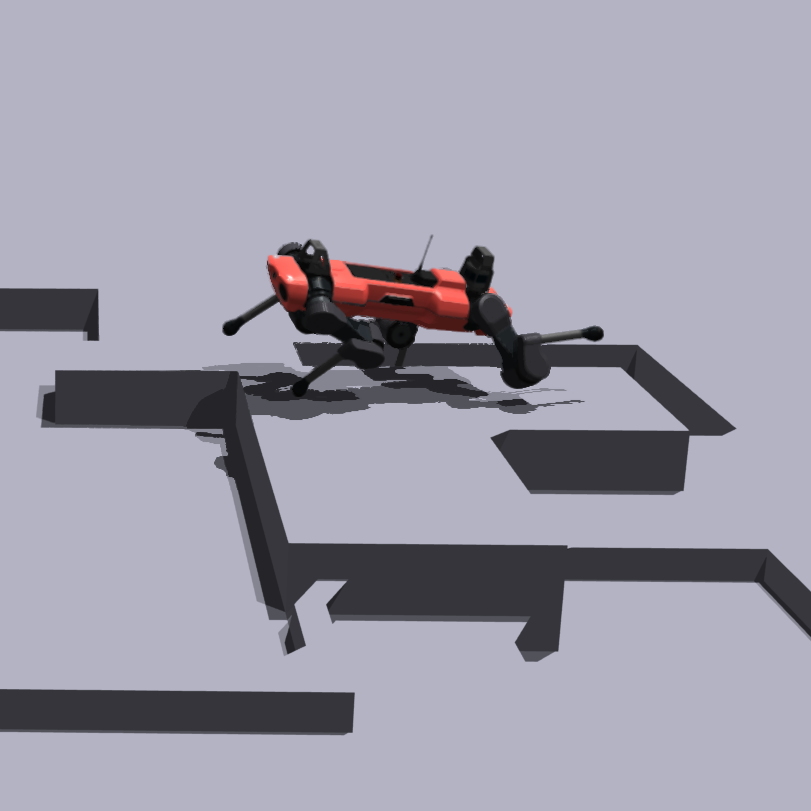}
     \end{subfigure}
     
     \vspace{2mm}
     
    \begin{subfigure}[b]{0.32\columnwidth}
        \centering
        \includegraphics[width=\textwidth, trim={0mm, 0, 0mm, 0}, clip]{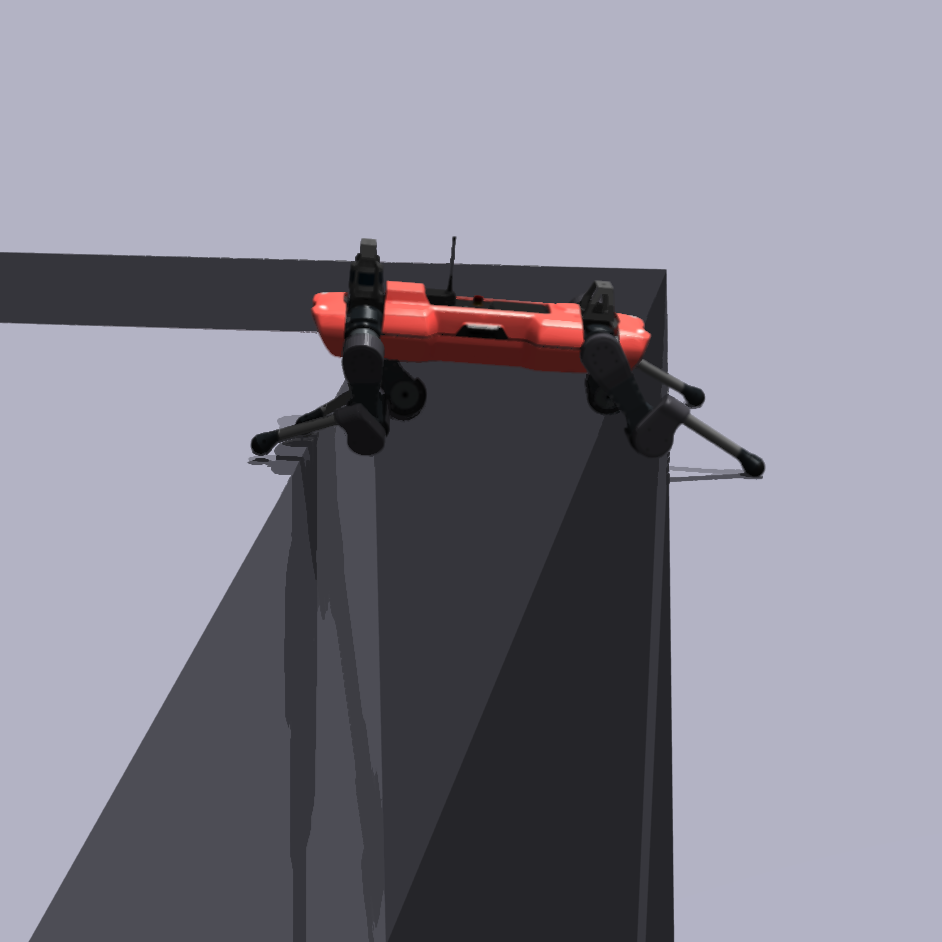}
     \end{subfigure}
     \hfill    
     \begin{subfigure}[b]{0.32\columnwidth}
     \centering
        \includegraphics[width=\textwidth, trim={0mm, 0, 0mm, 0}, clip]{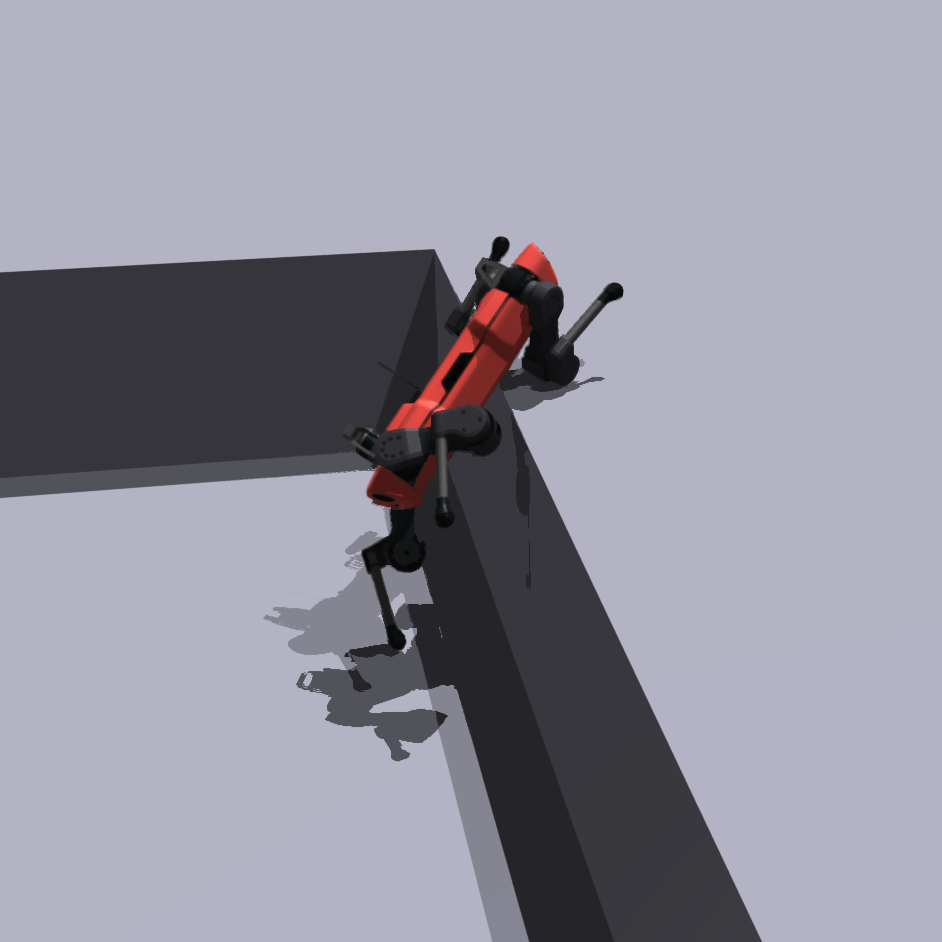}
     \end{subfigure}
     \hfill
    \begin{subfigure}[b]{0.32\columnwidth}
    \centering
        \includegraphics[width=\textwidth, trim={0mm, 0, 0mm, 0}, clip]{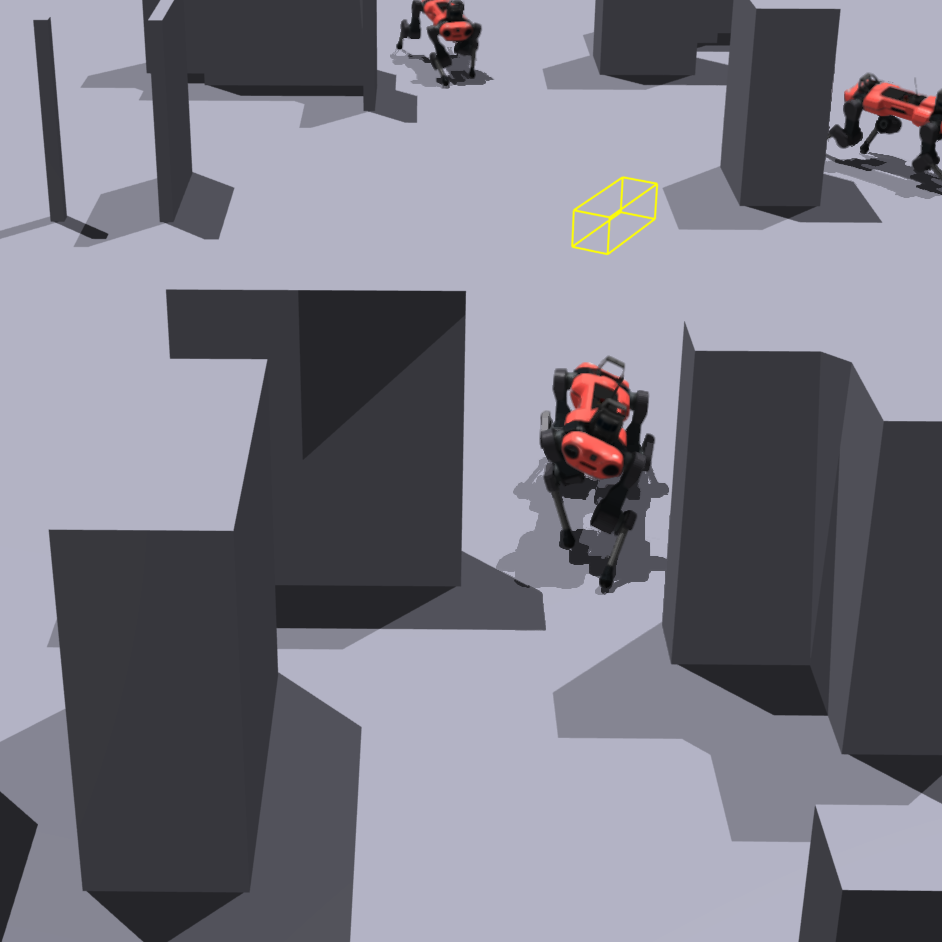}
     \end{subfigure}
    \caption{Simulation deployment of policies for each terrain type. The terrains represent the highest difficulty successfully solved by our approach with a \SI{95}{\%} success rate. Parameters of these terrains are provided in Table \ref{table:max_difficulty}.}
    \label{fig:sim_max_difficulty}
\end{figure}

\begin{table}[tb!]\centering
\caption{Maximum difficulty of terrains solved with a \SI{95}{\%} success rate. $V$: velocity tracking, $P_c$: continuous position tracking, $P_f$: final position tracking (ours).}
\begin{tabular}{rc|ccc}
    Terrain type & Parameter & \multicolumn{3}{c}{\SI{95}{\%} Success} \\
    & & V & $P_c$ & $P_f$ (ours)\\
    \toprule
     stairs (1) & rise & \SI{0.22}{m} & \SI{0.2}{m} & \SI{0.4}{m} \\
     slope (2) & angle & \SI{35}{\deg} & \SI{21}{\deg} & \SI{48}{\deg} \\
     random steps (3) & height & \SI{\pm0.14}{m} & \SI{\pm0.16}{m} & \SI{\pm0.2}{m}\\
     obstacles (4) & length & \SI{0.35}{m} & -  & \SI{0.85}{m}\\
     gap (5) & width & \SI{0.15}{m} & \SI{0.65}{m} & \SI{1.2}{m}\\
     pit (6) & depth & \SI{0.1}{m} & \SI{0.65}{m} & \SI{0.95}{m} \\
 \bottomrule
\end{tabular}

\label{table:max_difficulty}
\end{table}
We compare the capabilities of the three approaches by evaluating the maximum difficulty of terrains each policy can solve. We train policies on a single terrain type with an increased number of terrain levels and higher maximum complexity. After training for the same number of iterations, we evaluate the policies on each terrain difficulty and find the highest level they are able to solve.
The robots must cross \SI{4}{m} of the terrain within \SI{8}{s}. The velocity tracking policy receives a constant forward velocity command of \SI{1}{m/s}, and we consider a terrain solved if the robot has walked more than \SI{4}{m} even if it crashed afterwards.
For our approach, the policy receives a target \SI{5}{m} in front of its initial position, and we consider an episode successful if the final position of the robot is within \SI{0.5}{m} of the target and no crash has occurred. Even with a setup favoring velocity tracking, our approach outperforms both methods. In Table \ref{table:max_difficulty}, we report the highest terrain difficulty on which the different policies have a success rate of at least \SI{95}{\%}. The corresponding terrains are shown in Fig. \ref{fig:sim_max_difficulty}

Our approach raises the achieved difficulty on all terrains compared to velocity tracking. On stairs and slopes, it increases the step height and slope angle by \SI{80}{\%} and \SI{35}{\%}, which pushes the capabilities of the robot beyond what it would encounter in standard real-world conditions. Random rectangular obstacles remain a challenge due to the poor configuration of some of the generated terrains. Nevertheless, our approach increases the maximum obstacle height by \SI{40}{\%}. Finally, the most significant difference appears in our newly added terrains. The velocity tracking policy can only solve trivial terrain difficulties requiring minor adjustments to a standard walking gait. Our approach helps the policy learn more complex behaviors allowing the robot to perform jumps, climb using its knees and avoid obstacles of non-trivial size. Quantitatively, the size of obstacles, length of gaps, and depth of pits are increased by \SI{140}{\%}, \SI{700}{\%}, and \SI{850}{\%}, respectively.

On most terrains, the continuous position tracking approach achieves higher difficulties than velocity tracking. However, it completely fails to navigate obstacles with a sufficient reliability, and is inferior to final position tracking on all terrains.

\subsection{Energy Consumption}
\label{sec:energy}
\begin{figure}[!tb]
    \centering
    \includegraphics[width=0.9\columnwidth]{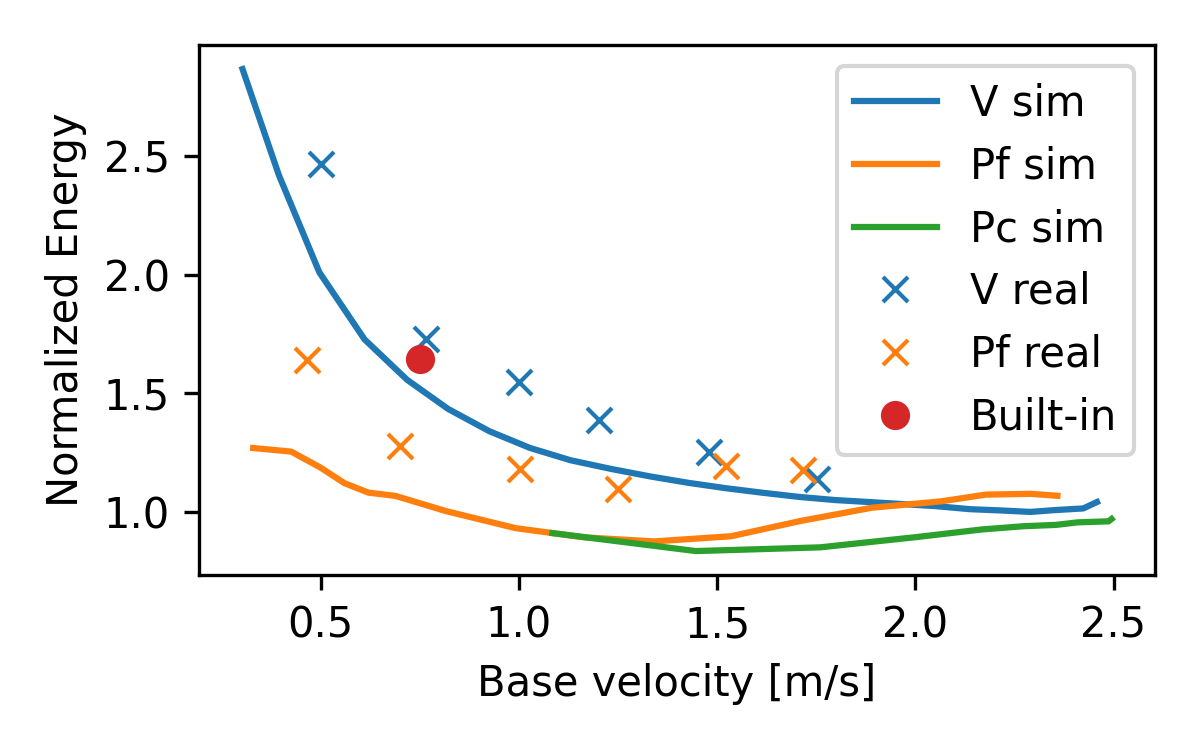}
    \caption{Normalized cost of transport of three approaches in simulation and on hardware. The sum of squared torques is recorded while the policies receive a constant command over \SI{100}{m} in simulation and \SI{5}{m} on the real robot. $V$: velocity tracking, $P_c$: continuous position tracking, $P_f$: final position tracking (ours). Built-in: learning based controller provided by the manufacturer.}
    \label{fig:cot}
\end{figure}
The policy does not need to compromise between maximizing task rewards and minimizing penalties with our approach. Furthermore, the robot's gait is less restricted than with the velocity tracking approach. In order to verify this hypothesis, we evaluate the energy consumption of policies trained with various approaches. We use the sum of squared torques as a proxy for energy since it is the penalized quantity during training and is proportional to heat losses in the motors.

We record the total energy consumption while the robots walk across \SI{100}{m} of flat terrain at various velocities. While it is straightforward to command the velocity tracking policy, our approach requires some modification to its commands. We can achieve a pseudo velocity control by fixing the target to be always \SI{3}{m} in front of the robot and setting a constant value instead of the time left to reach that target. By providing a fixed target and time, the policy can walk continuously even though it was never trained in that scenario. Furthermore, varying this fixed time directly influences the velocity of the robot. When the policy believes it has a lot of time to reach the target, it will adopt a slow and safe gait. As that time is reduced, it will try to close the gap faster and faster. We also modify the commands of the continuous position tracking policy. This time we change the distance to the target since the policy is not conditioned on time.

In Fig. \ref{fig:cot} we report the sum of squared torque per traveled distance for all approaches at different velocities. Interestingly, all approaches converge to similar values at high velocities, but the velocity tracking policy performs worse at slow speed. In that regime, keeping the base at constant velocity seems to have a stronger impact on the energy consumption. 
Both position based approaches present similar results, with the continuous tracking outperforming our approach at higher velocities. We note that this difference is expected since the continuous tracking policy is trained to run as fast as possible, while our approach is mostly trained with velocities around \SI{1}{m/s} (\SI{5}{m} in \SI{5}{s}).  

We see higher torques in hardware experiments compared to our simulated results. The offset can explained by the restrictive limits we impose on our actuators in simulation. The real robot is able to produce more torque than what the policy experienced in simulation. Nevertheless, we see similar trend, with our approach outperforming the velocity tracking policy at low speed and converging to similar torques at higher velocities. We do not transfer the continuous position tracking policy, since it is prone to aggressive motions which could potentially damage the robot.

\subsection{Emergent Gaits}
\begin{figure}[!tb]
    \centering
    \begin{subfigure}[b]{0.3\columnwidth}
        \centering
        \includegraphics[width=\textwidth, trim={3mm, 0, 3mm, 0}, clip]{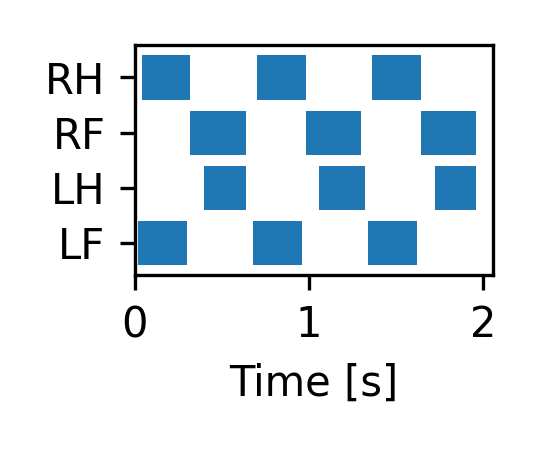}
        \caption{$V$}
     \end{subfigure}
     \hfill    
     \begin{subfigure}[b]{0.3\columnwidth}
     \centering
        \includegraphics[width=\textwidth, trim={3mm, 0, 3mm, 0}, clip]{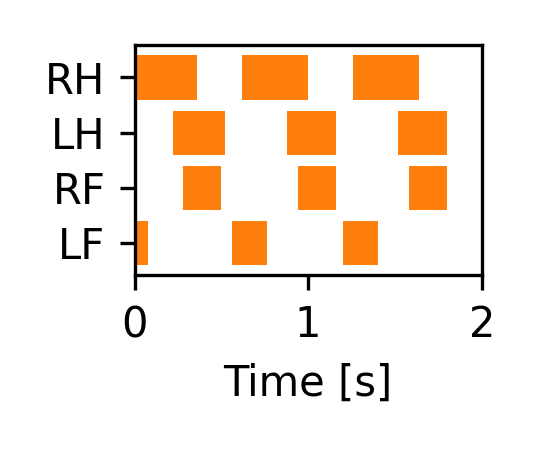}
        \caption{$P_f$}
     \end{subfigure}
     \hfill
    \begin{subfigure}[b]{0.3\columnwidth}
    \centering
        \includegraphics[width=\textwidth, trim={3mm, 0, 3mm, 0}, clip]{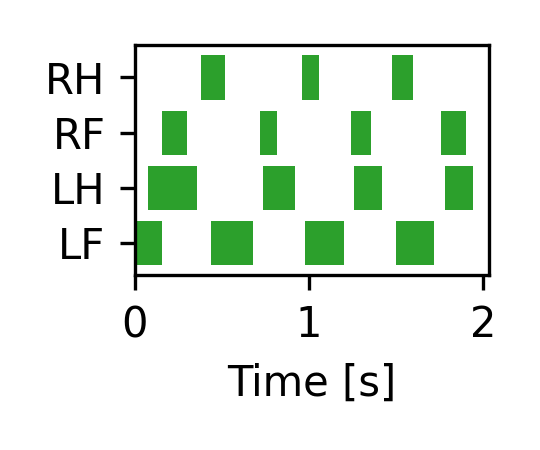}
        \caption{$P_c$}
     \end{subfigure}
    \caption{Emerging gait patterns of three task definitions. (a): Velocity tracking, trotting gait. (b): Final position tracking (ours), three-phased gait towards the front-left side. (c): Continuous position tracking, three-phased gait towards the right-hind side.}
    \label{fig:gaits}
    \vspace{-5mm}
\end{figure}

\begin{figure*}[!t]
    \centering
    \begin{subfigure}{\textwidth}
    \centering
        \includegraphics[width=0.45\textwidth, trim={3mm, 1.25mm, 3mm, 2mm}, clip]{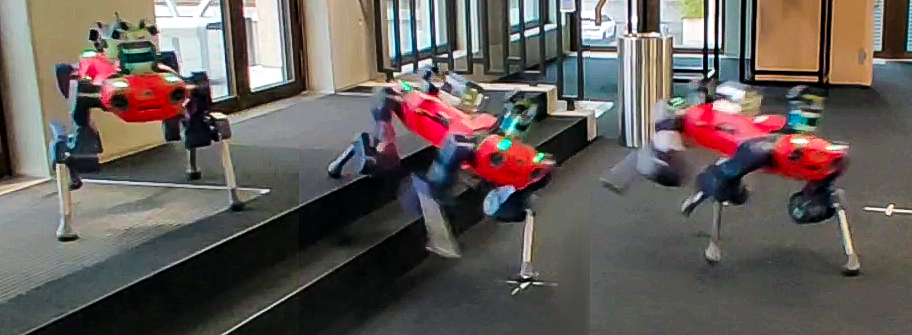}
        \includegraphics[width=0.45\textwidth, trim={0mm, 70mm, 0mm, 70mm}, clip]{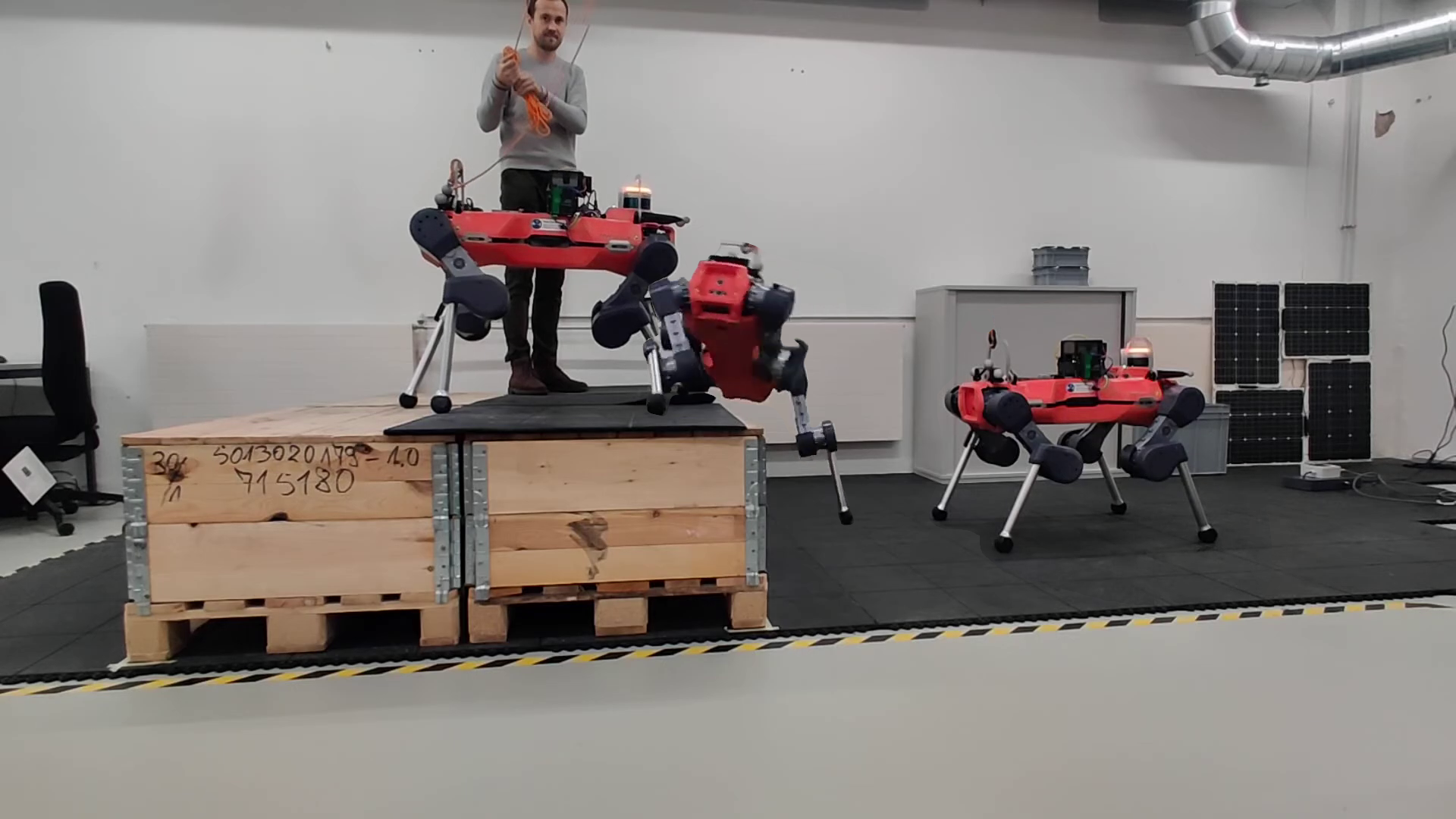}
    \end{subfigure}
    \vspace{1mm}
    \begin{subfigure}{\textwidth}
    \centering
        \includegraphics[width=0.45\textwidth, trim={3mm, 0, 3mm, 1mm}, clip]{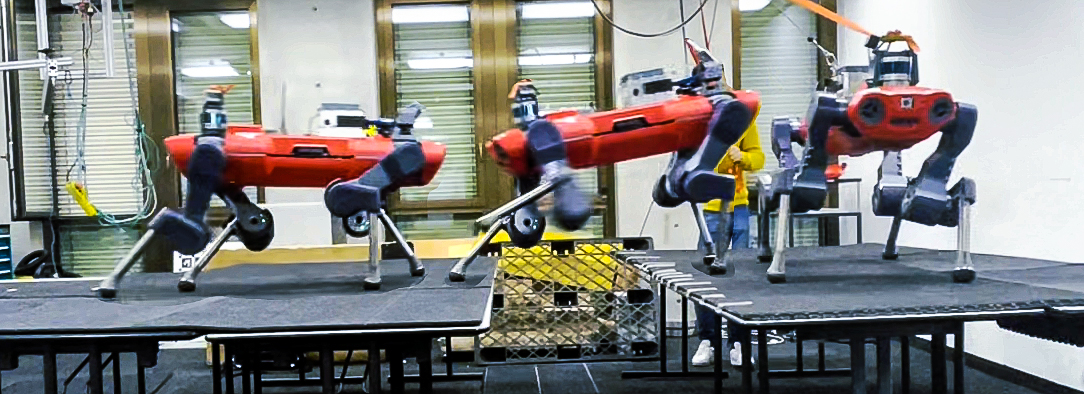}
        \includegraphics[width=0.45\textwidth, trim={0mm, 10mm, 0mm, 3mm}, clip]{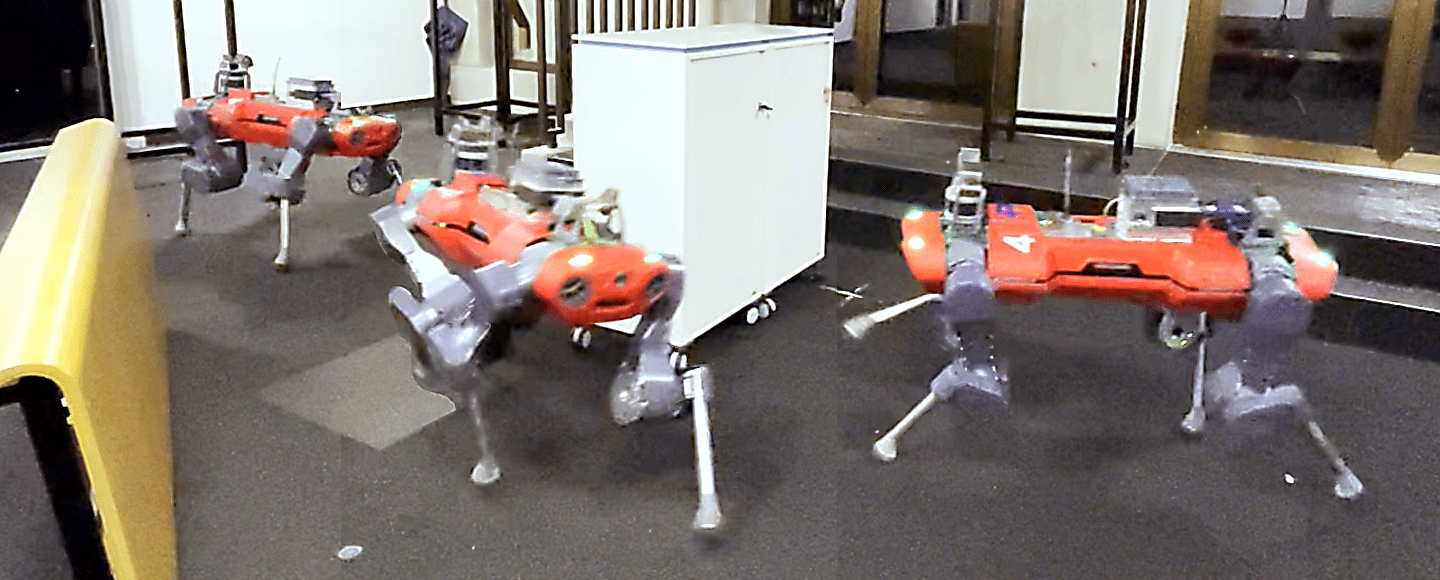}
    \end{subfigure}
    
    \caption{Deployment of policies on the real robot. Top Left: running on stairs, top right: climbing on a \SI{0.55}{m} box, bottom left: jumping over a \SI{0.6}{m} gap, bottom right: navigating around obstacles.}
    \label{fig:real}
\end{figure*}

Most previous works on learned locomotion resulted in a trotting gait. This can be explained by the fact that the robot's base can keep a nearly constant velocity. As such, velocity tracking approaches are heavily biased towards trotting. However, walking with a constant base velocity is not common for animals as it requires additional effort. With our approach, we remove these artificial constraints, which allow the emergence of different types of gaits. Furthermore, for velocity tracking policies, it is necessary to encourage larger steps by rewarding the air time or clearance of the feet \cite{JeminAnymal, RudinMinutes}. This is needed because the policy prefers to do many small steps in order to follow the command with minimal oscillations of the base. With our approach, this reward is no longer necessary since base oscillations are no longer penalized.

Fig. \ref{fig:gaits} compares the gaits of velocity and position tracking. We see that the policy learns a three-phased asymmetric gait with our approach. Interestingly, the robot moves diagonally, probably due to the kinematics and motor arrangement, giving the most extended reach and maximum power in this direction. In Fig. \ref{fig:gaits}, the robot is moving towards the front left. The gait is split into three phases. First, the back (right hind) leg touches the ground, followed by the center (left hind and right front) legs. Finally, the forward facing (left front) leg is briefly in contact. The continuous tracking policy uses a similar gait, but it learns to walk towards it's right back side, meaning that the contact schedule is inverted.

Qualitatively, as can be seen in the accompanying video, the robot's motion is more dynamic and is generally perceived to be more natural with our approach. The combination of a stronger base motion and large smoother steps gives the robot an organic appearance, which the velocity tracking policy lacks.

However, there is an unfortunate artifact in the behavior. The policy only learns to walk in one direction, preferring to turn instead of walking sideways or backward. This can become a problem when we want the robot to walk back and forth repetitively. This seems to be due to a very attractive local minimum as every training run results in a random but fixed walking direction.

\subsection{Hardware Deployment}

We deploy policies trained with our approach on the real robot. Following previous work \cite{JeminAnymal}, we augment the simulation with a neural network trained to mimic the series elastic actuators of the robot. Unfortunately, the actuator network is not accurate enough in the high-velocity regime. At maximum velocity, it still allows high torques, which is not possible for a brushless DC motor \cite{DcMotor}. We model the maximum torque as an affine function of motor velocity and clip the torques required by the policy to those maximum values.

During deployment, commands can be specified in two ways: either with real position targets or using a joystick with the pseudo velocity control described in Sec. \ref{sec:energy}. In the second case, the direction of the joystick command defines the target direction, while the magnitude is used to scale the remaining time given to the policy. This proves to be a natural way to command the robot, and the policy responds well to quickly changing commands even though it was not trained for it.

For practical reasons, we do not use the most complicated terrains described in Table \ref{table:max_difficulty}. Nevertheless, we show that our approach allows the robot to perform maneuvers that were previously not possible. We show our policies successfully running on flat terrain, climbing stairs at high speed, jumping over a \SI{0.6}{m} gap, climbing on a \SI{0.55}{m} box, and navigating around obstacles. These motions are represented in Fig. \ref{fig:real} as well as in the supplementary video.

The success rate of deployed policies greatly depends on the terrain type. While it is similar to \ref{table:max_difficulty} on stairs, it is lower for harder terrains. This is due to our reliance on provided state estimation and perception modules which were not designed for jumping or climbing. A motion capture system was required to successfully deploy the climbing policy.

\section{Discussion}
In this work, we have proposed to rethink the task definition used to train locomotion policies for legged robots. We have shown that specifying a target location instead of a base velocity command reduces the constraints on the solution space. This set-up decouples the task from the optimization of other quantities, lets the policy learn a more energy-efficient gait, and most importantly, helps it learn more complex maneuvers. On the downside, our approach seems more prone to training instabilities. Another challenge is that the robot only learns to walk in one direction and does not use its kinematic symmetry. The task should also be extended to allow the specification of multiple consecutive way-points without discontinuities in the motion. Finally, the reliance on state estimation and perception has to be reduced or these modules need to be improved for our specific use cases. Solving these challenges should be part of future work.

Our approach can be applied to a large set of tasks and robots. It is not limited to locomotion on challenging terrain. For example in a fall recovery scenario, the reward encouraging the robot to stand up can also be activated only at the end of an episode, letting the policy free to choose the trajectory instead of requiring it to recover as fast as possible.

By showing that these relatively small changes in the training process lead to great improvements in the agility of the robot, we emphasizes that we are still far from exploiting the full capabilities of our robots, and encourage other researchers to apply similar techniques to their respective tasks.

\addtolength{\textheight}{-5cm}   









\bibliographystyle{IEEEtran}
\bibliography{IEEEabrv, bibliography}

\end{document}